\pdfoutput=1

\documentclass[11pt]{article}

\usepackage{acl}
\usepackage{times}
\usepackage{colortbl} 
\usepackage{booktabs} 
\usepackage{tabularx}
\usepackage{fontawesome}
\usepackage{listings}
\usepackage{latexsym}
\usepackage{enumitem}
\usepackage{floatrow} 
\usepackage{booktabs}
\usepackage{longtable} 
\usepackage{colortbl} 
\usepackage{adjustbox} 
\usepackage{array} 
\usepackage{amsmath}  
\usepackage{cleveref}
\usepackage{adjustbox}
\usepackage{graphicx}    
\usepackage{caption}     
\usepackage{subcaption}  
\usepackage{afterpage}   
\usepackage{longtable}
\usepackage{makecell}
\usepackage{natbib}
\usepackage{multirow}
\usepackage{todonotes}
\usepackage[T1]{fontenc}

\usepackage[utf8]{inputenc}

\usepackage{microtype}

\usepackage{inconsolata}

\usepackage{graphicx}
\title{Shaping the Safety Boundaries: Understanding and Defending Against Jailbreaks in Large Language Models

}

\author{
\textbf{Lang Gao}$^{1,2}$,~~\textbf{Jiahui Geng}$^{1}$,~~\textbf{Xiangliang Zhang}$^{3}$,  \textbf{Preslav Nakov}$^{1}$,   \textbf{Xiuying Chen}$^{1}$\thanks{Corresponding author.} \\
\\
$^{1}$MBZUAI~  $^{2}$Huazhong University of Science and Technology~\\ $^{3}$University of Notre Dame\\
\texttt{\{Lang.Gao, Jiahui.Geng, Preslav.Nakov, Xiuying.Chen\}@mbzuai.ac.ae}\\
\texttt{xzhang33@nd.edu}
}

\begin{document}
\maketitle
\begin{center}
    \vspace{10pt}
    \small
    \textcolor{orange}{\bf \faWarning\, WARNING: This paper contains model outputs that may be considered offensive.}
\end{center}
\renewcommand{\thefootnote}{\fnsymbol{footnote}}
\raggedbottom
\renewcommand{\textcolor}[2]{#2} 

\begin{abstract}
Jailbreaking in large language models (LLMs) poses major security risks by tricking models into generating harmful text. 
However, there is limited understanding of how jailbreaking operates, which makes it difficult to develop effective defenses.
\textcolor{blue}{
In this work, we conduct a large-scale analysis of seven jailbreak methods and uncover that inconsistencies in previous studies arise from insufficient observation samples.
Our analysis reveals that jailbreaks shift harmful activations beyond a defined \textit{safety boundary}, where LLMs become less sensitive to harmful information. 
We also find that the low and the middle layers are critical in driving these shifts, while deeper layers play a lesser role.}
Leveraging these insights, we propose a novel defense mechanism called \textit{Activation Boundary Defense} (ABD), which adaptively constrains activations within the safety boundary. 
\textcolor{blue}{To further optimize performance, we use Bayesian optimization to select the most effective layers for ABD application and confirm that the low and middle layers have the greatest impact, consistent with our earlier observations.}
Experiments across multiple benchmarks demonstrate that ABD achieves an average defense success rate (DSR) of over 98\% against various jailbreak attacks, with less than 2\% impact on the model's overall capabilities.
\end{abstract}

\section{Introduction}

The widespread use of Large Language Models (LLMs) across various fields~\cite{kaddour2023challenges,chen2025unveiling,xie2025medtrinitym,liu2024vuldetectbenchevaluatingdeepcapability} has raised concerns about the safety of the output and the robustness of these models~\cite{ye2024justice,huang2025trustworthiness,wang2025trusteval}. 
It has been shown that \textit{jailbreak attacks}, which use crafted prompts to deceive LLMs into generating harmful content, can bypass LLM's safety alignment~\cite{liu2024autodan,zou2023universal,song2025injecting},
and a lot of research has focused on developing defense mechanisms and counter-prompts to mitigate such attacks~\cite{robey2023smoothllm,xie-etal-2024-gradsafe}.

\begin{figure*}[t]
\centering
\includegraphics[scale=0.21]{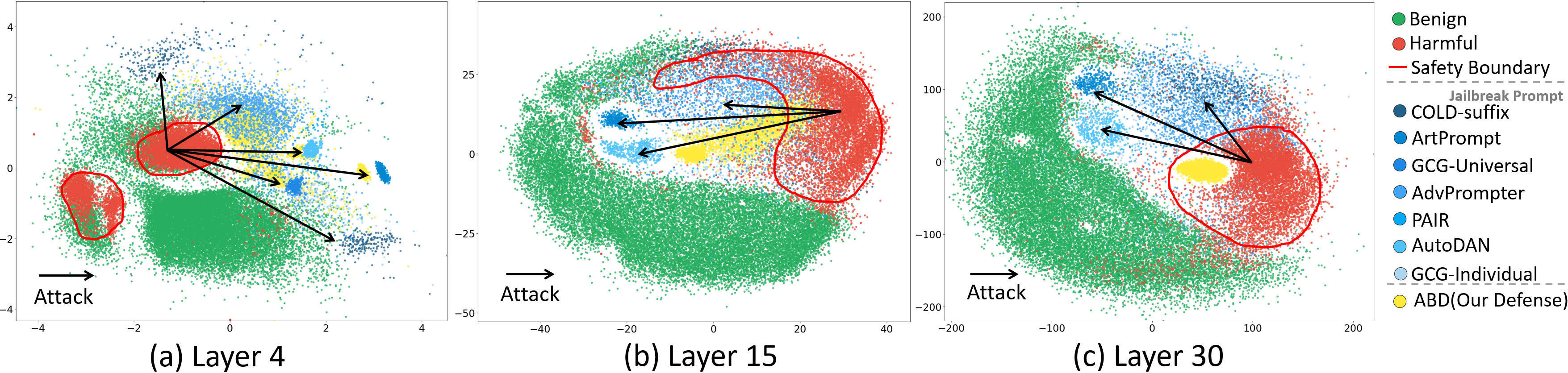}
\caption{
Projected activation space overview of Vicuna-7B-v1.3 across different layers.
Harmful activations are observed to cluster together, and we define the surrounding boundary as the \emph{safety boundary}.
The attack arrow indicates that jailbreak prompts shift harmful activations into the benign space to evade safety checks.
}
\label{fig:intro}
\end{figure*}

Understanding the internal mechanisms of jailbreak attacks is crucial for both explaining their operation and developing effective defenses.
\textcolor{blue}{While existing studies have made initial explorations \cite{yu2023jailbreak,ball2024understanding,lin-etal-2024-towards-understanding}, fundamental disagreements remain about the attack mechanisms.
The academic community continues to debate two key aspects of explainability: (\emph{i})~The localization of attack impacts across the model layers, with competing findings suggesting primary responsibility in low \cite{he2024jailbreaklensinterpretingjailbreakmechanism}, middle \cite{zhou2024alignment,shen2024jailbreak}, or deep layers \cite{li2024rethinking}, and (\emph{ii})~the progression pattern of malicious activation, with conflicting observations about gradual shifts versus abrupt transitions in the middle layers \cite{zhao2024eeg,shen2024jailbreak}. 
Our experiments suggest that these inconsistencies may stem from methodological limitations in prior analysis, particularly their reliance on small prompt samples (typically \(\sim 100\)).
On the defense front, current approaches face dual challenges: Most detection methods either require additional training \cite{zhao2024eeg} or demonstrate limited generalizability due to small-sample probing \cite{shen2024jailbreak}, while mitigation strategies often degrade normal model capabilities. }

\textcolor{blue}{In this work, we aim to better understand the mechanism of how jailbreaking works.
In particular, we provide our explanation of jailbreaks based on a comprehensive analysis of over \textit{30,000} samples, a significantly larger scale than previous studies. 
Figure~\ref{fig:intro} shows the projection of three types of prompts: \textit{benign prompts} that contain no harmful information, \textit{harmful prompts} that attempt to induce the LLM to generate harmful content but fail, and various \textit{jailbreak prompts} that successfully induce harmful outputs, collected from different types of jailbreak attack methods.
The projection is based on the last token representations across different layers, referred to as \textit{activations}, as they capture the model's overall understanding of the entire input sequence~\cite{radford2019language,zou2023universal}. 
Our results reveal distinct clustering patterns, where different prompts form separate activation clusters. 
Notably, harmful prompts form a contained cluster within a specific region that we define as the \textit{safety boundary}, representing the LLM's internal safety check that typically prevents harmful activations from escalating into harmful outputs.
In contrast, jailbreak prompts push activations beyond the safety boundary into an unmonitored space where harmful content can be generated. This shift is most pronounced in the low and middle layers, indicating their critical role for jailbreak success.
}

This finding motivates a new defense mechanism that we propose, \textit{Activation Boundary Defense} (ABD), which constrains jailbreak prompt activations within a safety boundary by applying penalties to activation values. 
Specifically, we impose minimal penalties within the boundary, but sharply increase them beyond it, thereby preserving model utility.
The penalty function considers various factors, including the defense layer and the defense dimension, which balance the performance on defense and on general tasks. 
We propose a novel objective that maximizes the Defense Success Rate (DSR) while minimizing the number of penalized layers.
This objective is implemented with an adapted Bayesian Optimization method~\cite{jones1998efficient}, which iteratively proposes and refines selection suggestions.

Experiments on AdvBench \cite{zou2023universal} show that ABD has good generalization, achieving an average DSR of over 98\% against various forms of jailbreak attacks. 
For general ability tasks, the performance drop of the model equipped with ABD does not exceed 2\%, which is substantially lower than the drop of up to 37\% with other defenses.

Our contributions can be summarized as follows:

\noindent $\bullet$ We uncover a comprehensive activation distribution and introduce the \emph{safety boundary}, resolving contradictions in prior work and highlighting the vulnerability of the low and middle layers.

\noindent $\bullet$  We propose a lightweight, extensible defense that penalizes only targeted samples and a few key layers, ensuring efficiency and precision.

\noindent $\bullet$  Our experiments on benchmark datasets demonstrate that ABD achieves an average DSR of over 98\% against various jailbreak attacks, with less than a 2\% impact on general capabilities.

\section{Related Work}

\paragraph{Jailbreak attacks on LLMs.}  
Jailbreak attacks craft prompts to induce harmful outputs from LLMs, evolving from early manually designed approaches to more sophisticated techniques. Early methods, such as DAN \cite{shen2023anything}, rely on manually constructed prompts. Later, optimization-based methods \cite{zou2023universal,liu2024autodan,guo2024cold} emerged, using iterative optimization strategies to refine harmful prompts and enhance their stealthiness. \textcolor{blue}{Model-based methods \cite{chao2023jailbreaking,ding-etal-2024-wolf,paulus2024advprompter,huang2024obscureprompt} use attacker LLMs to autonomously generate and improve jailbreak prompts.}
Meanwhile, rule-based methods \cite{jiang-etal-2024-artprompt,liu2024flipattackjailbreakllmsflipping} rewrite prompts using predefined rules that effectively deceive the model. 
In this work, we analyze the mechanisms behind these jailbreak techniques.

\paragraph{Defense against jailbreak attacks.}
\textcolor{blue}{Defense strategies for jailbreak attacks typically fall into three categories. One common approach enhances model safety and robustness by reformulating inputs to counter jailbreak prompts through techniques such as backtranslation~\cite{wang2024defendingllmsjailbreakingattacks}, paraphrasing~\cite{jain2023baseline}, self-reminding~\cite{xie2023defending}, and adversarial prompting~\cite{zhou2024defending}. Another approach focuses on identifying jailbreak prompts using indicators, such as classifying input sequences based on perplexity and sequence length~\cite{alon2023detecting} or using linear classifiers to detect jailbreak activations and decide whether to respond~\cite{zhao2024eeg}.
A more recent trend directly manipulates the model's internal representations~\cite{xu-etal-2024-safedecoding,li2024rethinking,liu-etal-2024-alignment} or editing the model~\cite{zhao2024defending}.}
Unlike these methods, our approach directly constrains activations within a safety boundary, avoiding extra tokens or modules.

\paragraph{Mechanistic interpretability of LLMs.}

The growing concern about LLM safety has sparked increasing interest in interpreting LLM features in jailbreak prompts.
For example, \citet{ball2024understandingjailbreaksuccessstudy} identified a common mechanism whereby jailbreaks reduce the harmfulness perception in most LLMs. Similarly, \citet{li2024rethinking} investigated patterns that trigger the model to recognize safety issues.
\citet{zhou2024alignment} discovered that the vocabulary mappings of activations significantly changed when processing jailbreak inputs.
Research efforts have also focused on proposing corresponding defense methods based on interpretability results.
For instance, \citet{zhao2024eeg} and \citet{shen2024jailbreak} modeled how jailbreak activations transfer between benign and harmful activation spaces as layers deepen. They designed adaptive defenses whose strength varies across different layers.
\textcolor{blue}{However, the limited data in all previous studies often led to controversy and ambiguity.} 

\section{Understanding Jailbreaks in LLMs}\label{jailbreak_mechanisms}
\subsection{Controversy in Literature}
\begin{figure}[t]
\centering   
    \includegraphics[width=0.9\textwidth]{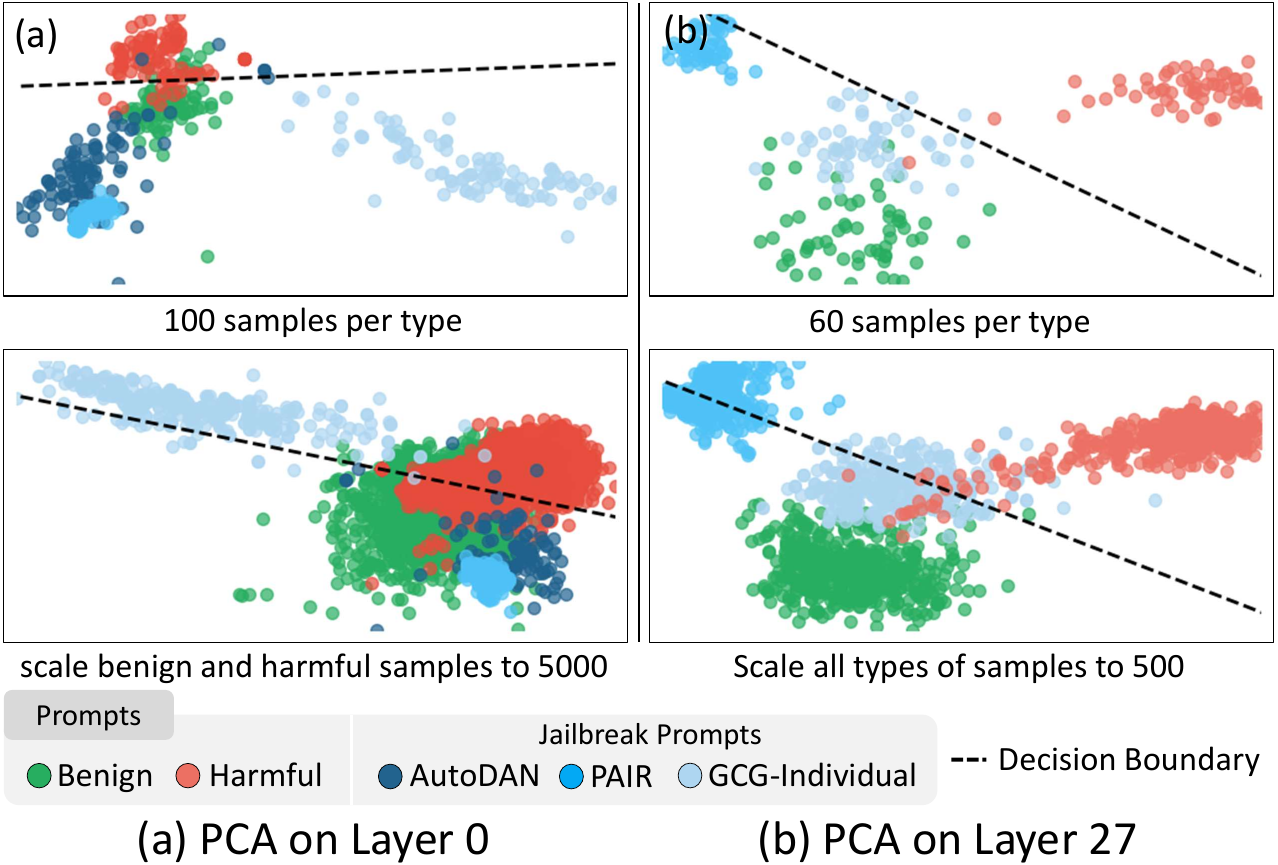}
    \caption{
PCA visualizations reveal the limitations of small-sample sizes:  
(a) Top: 100 samples per type~\cite{he2024jailbreaklensinterpretingjailbreakmechanism}; Bottom: 5,000 benign and harmful samples.  
(b) Top: 60 samples per type~\cite{zhao2024eeg}; Bottom: 500 samples per type.
    }
    \label{fig:sample_affect}
\end{figure}

\paragraph{\textcolor{blue}{Disputes on layer importance.}}\label{understand_key_layers}
\textcolor{blue}{There have been different opinions about which layers are most important, mainly due to varying perspectives. 
For example, \citet{he2024jailbreaklensinterpretingjailbreakmechanism} argued that the low layers were essential because they treated jailbreak activations as benign ones.
\citet{zhou2024alignment} believed that the low and the middle layers were essential based on tokens generated from activations. 
\citet{li2024rethinking} argued that deep layers were critical, as their proposed safety pattern showed greater values in these layers.
Similarly,~\citet{he2024jailbreaklensinterpretingjailbreakmechanism} focused on activations, as we do as well.
However, their experimental setup suffers from a key limitation: a small sample size used in the analysis. 
As shown in Figure~\ref{fig:sample_affect}(a), under the original settings, benign and harmful activations are linearly separable, with most jailbreak samples misclassified as benign. After scaling, half of the jailbreak activations align with harmful ones, making the separation nonlinear. This underscores the risks of small-scale studies and the need for large-scale analysis.}

\paragraph{Disputes on the jailbreak mechanism.} \label{understand_jailbreak}
There are differing perspectives on the internal mechanism of jailbreaks.
\citet{zhao2024eeg} viewed jailbreak as a gradual process, where activations transition from harmful to benign spaces as they pass from the low to the deep layers. 
In contrast, \citet{shen2024jailbreak} argued that jailbreaks manifest abruptly in the deeper layers, with activations positioned between harmful and benign spaces.
\textcolor{blue}{Both studies relied on fewer than 100 samples per activation type, contributing to similar inconsistencies. 
As shown in Figure~\ref{fig:sample_affect}(b), before scaling, jailbreak activations align closely with benign ones, consistent with \citet{zhao2024eeg}. 
However, after scaling, they diverge from both harmful and benign activations.
When considering \citet{shen2024jailbreak}, similar conflicts arise between pre- and post-scaling results (details in Appendix~\ref{tsne-exp}, Figure~\ref{fig:sample_affect_tsne}). 
These conflicting observations cast doubt on their conclusions regarding the true mechanism of jailbreaks.}

\subsection{Experimental Settings}\label{concepts}

To address the above limitations, we conducted a large-scale analysis on a 300 times greater dataset.

\begin{figure*}[t]
\centering   
    \includegraphics[width=\textwidth]{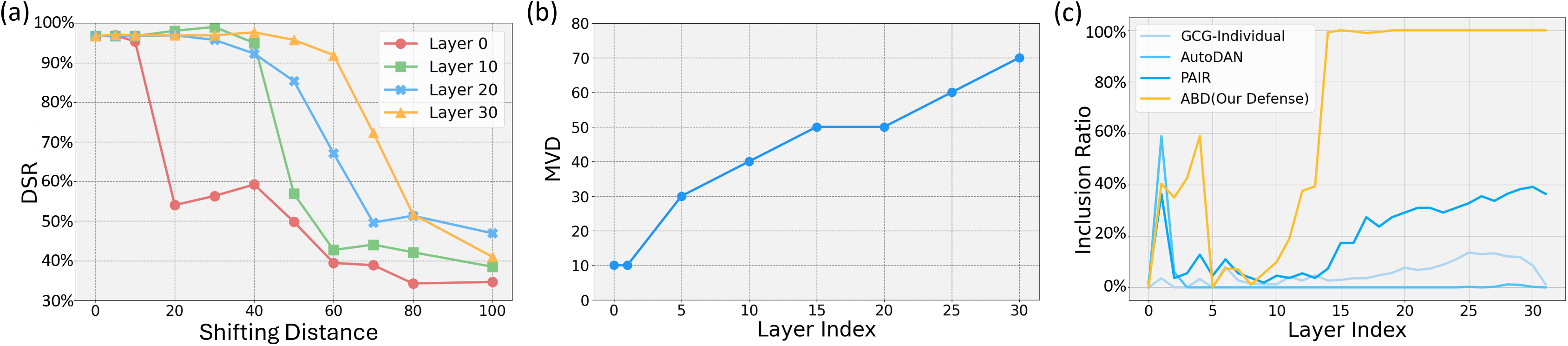}
    \caption{
     (a) Impact of random activation shifts across layers.
 DSR (Defense Success Rate) decreases as shifting distance increases, regardless of the affected layers.
(b) MVD (Most Vulnerable Distance) across layers.
MVD increases as layers go deeper.
(c) Inclusion ratio of jailbreaking activations in the harmful activation space.
Without ABD, the ratio stays below 0.4 but rises to 1 when ABD is applied.
    }
    \label{fig:act_shift}
   
\end{figure*} 

\paragraph{Dataset and Model.}
Our dataset consists of 32,507 samples in three categories: benign, harmful, and jailbreak.
For benign samples, we used a subset from the Alpaca dataset \cite{alpaca}, as it has been carefully curated to ensure that only safe content is included.
We collected harmful samples from five datasets in RedEval \cite{bhardwaj2023redteaming} and AdvBench \cite{zou2023universal}.
These datasets encompass a wide variety of harmful and misleading prompts.
For jailbreak samples, we applied seven different jailbreak attack methods, including GCG-Individual and GCG-Universal \cite{zou2023universal}, AdvPrompter \cite{paulus2024advprompter}, COLD-Suffix Attack \cite{guo2024cold}, AutoDAN \cite{liu2024autodan}, PAIR \cite{chao2023jailbreaking}, and ArtPrompt \cite{jiang-etal-2024-artprompt}, on AdvBench. Detailed statistics can be found in Table~\ref{tab:sample_statistics} in Appendix~\ref{stats}. 
We used Vicuna-7B-v1.3 \cite{vicuna2023}, a 32-layer Transformer, fine-tuned on a safety-aligned LLM.
Although it can identify harmful information, it remains vulnerable to jailbreak attacks~\cite{chu2024comprehensive}, making it an ideal model for analyzing attacks and defenses.

\paragraph{MDS projection.} 
Our projection was based on the \textit{activation}, the last token vector of the input. 
\textcolor{blue}{As discussed in \S\ref{understand_key_layers}, linear classification may be insufficient to model the boundary between different prompts. Therefore, we adopted Multi-Dimensional Scaling (MDS) \cite{carroll1998multidimensional} as our dimensionality reduction method, as PCA assumes linear separability~\cite{anowar2021conceptual}, and t-SNE focuses on preserving local structures while distorting the global distribution~\cite{van2008visualizing}.}
\subsection{Jailbreak Mechanisms Findings}

\subsubsection{Activation Distribution}\label{activation_space}  
We present the activation distributions of representative layers in Figure~\ref{fig:intro}, with the full version available in Figures~\ref{fig:all_subfig1},\ref{fig:all_subfig2},\ref{fig:all_subfig3} in Appendix~\ref{full_view}.
\textcolor{blue}{Our direct observations are as follows: } 
(1)~\textit{Harmful and benign activations overlap and are not linearly separable in most layers.} 
\textcolor{blue}{There are no clear linear decision boundaries to distinguish between harmful and benign activations, highlighting the ineffectiveness of treating jailbreak activations as a simple linear classification problem.}
(2)~\textit{Jailbreak activations shift significantly out of the harmful activation space}, 
\textcolor{blue}{forming distinct regions with minimal overlap. This shift is consistent across different jailbreaks, starting in the low layers and persisting throughout the model, rather than originating in the middle or deep layers~\cite{zhou2024alignment, shen2024jailbreak}. Additionally, the shift is most pronounced in the low and middle layers.}

\subsubsection{Safety Boundary}\label{safety_boundary}
\textcolor{blue}{Based on observations of jailbroken and harmful activations, we propose that jailbreak mechanisms are better explained by activations shifting outside a closed safety boundary, rather than simply crossing from one side of a linear boundary to the other. To validate this, we conduct experiments and introduce two key concepts below.} 
\textcolor{blue}{\textit{Randomized Activation Shifting (RAS)}: It is an approximation of jailbreak attacks by randomly shifting the original activation.
Assume the activation on the $l$-th layer is $a_l$, we shift it by a distance $r$ in a random direction $\hat{u}$:}
\begin{align}
    a_l \leftarrow a_l + r \cdot \hat{u}.
\end{align}
Figure~\ref{fig:act_shift}(a) shows the changes in Defense Success Rate (DSR) under different values of $r$ for four selected layers. 
\textcolor{blue}{The definition of DSR is provided in the Appendix~\ref{cal_dsr}.
Initially, at $r = 0$, the DSR exceeds 99\% as harmful activations should lie within the safety boundary. As $r$ increases, the DSR decreases across layers, with a sharp drop in a specific range (e.g., $r = 10$–$20$ on layer 0) and a more gradual decline elsewhere. \textcolor{blue}{The decrease indicates the weakening sensitivity of harmful contents, and the sharp drop suggests the presence of\textit{ an implicit safety boundary}.
Appendix~\ref{case-study} presents how model responses vary with different RAS levels in Table~\ref{tab:case-study}.}}

\textit{Most Vulnerable Distance (MVD)}: 
\textcolor{blue}{It is a measurable approximation of the safety boundary.}
MVD represents the specific distance at which the DSR experiences the sharpest drop, indicating the point where the model's ability to detect harmful activations becomes most vulnerable:
\begin{align}
    \mathrm{MVD = \textstyle \arg\min_r}\textstyle \frac{\mathrm{d(DSR)}}{\mathrm{d}r},
\end{align}
where $\frac{\mathrm{d(DSR)}}{\mathrm{d}r}$ denotes the rate of change of DSR with respect to  $r$. 
As shown in Figure~\ref{fig:act_shift}(b), 
\textcolor{blue}{MVD increases with model depth, indicating that the safety boundary expands in deeper layers and requires larger shifts to bypass safety checks. 
This aligns with Figure~\ref{fig:intro}: in lower layers (Figure~\ref{fig:intro}(a)), jailbreak activations scatter widely but surpass the safety boundary; 
in middle layers (Figure~\ref{fig:intro}(b)), jailbreak activations align more with benign activations while remaining outside the boundary; 
in deeper layers (Figure~\ref{fig:intro}(c)), jailbreak activations cluster, partially overlapping the safety boundary.}

\subsubsection{Key Takeaways on the Jailbreak Process} 

Based on the above analysis, we summarize the jailbreak process as follows:
(1)~Jailbreak takes effect in the low layers, contrary to prior work~\cite{zhou2024alignment}, which suggested it begins in the middle layers.
(2)~As the attack progresses, jailbreak continues to shift the activation outside the safety boundary, causing the model to be deceived.
(3)~When comparing all layers, the low and the middle layers exhibit the strongest shift and smallest safety boundaries, demonstrating their importance in the jailbreak process.


\section{Activation Boundary Defense}

\begin{figure}[t]
\centering   
    \includegraphics[width=\textwidth]{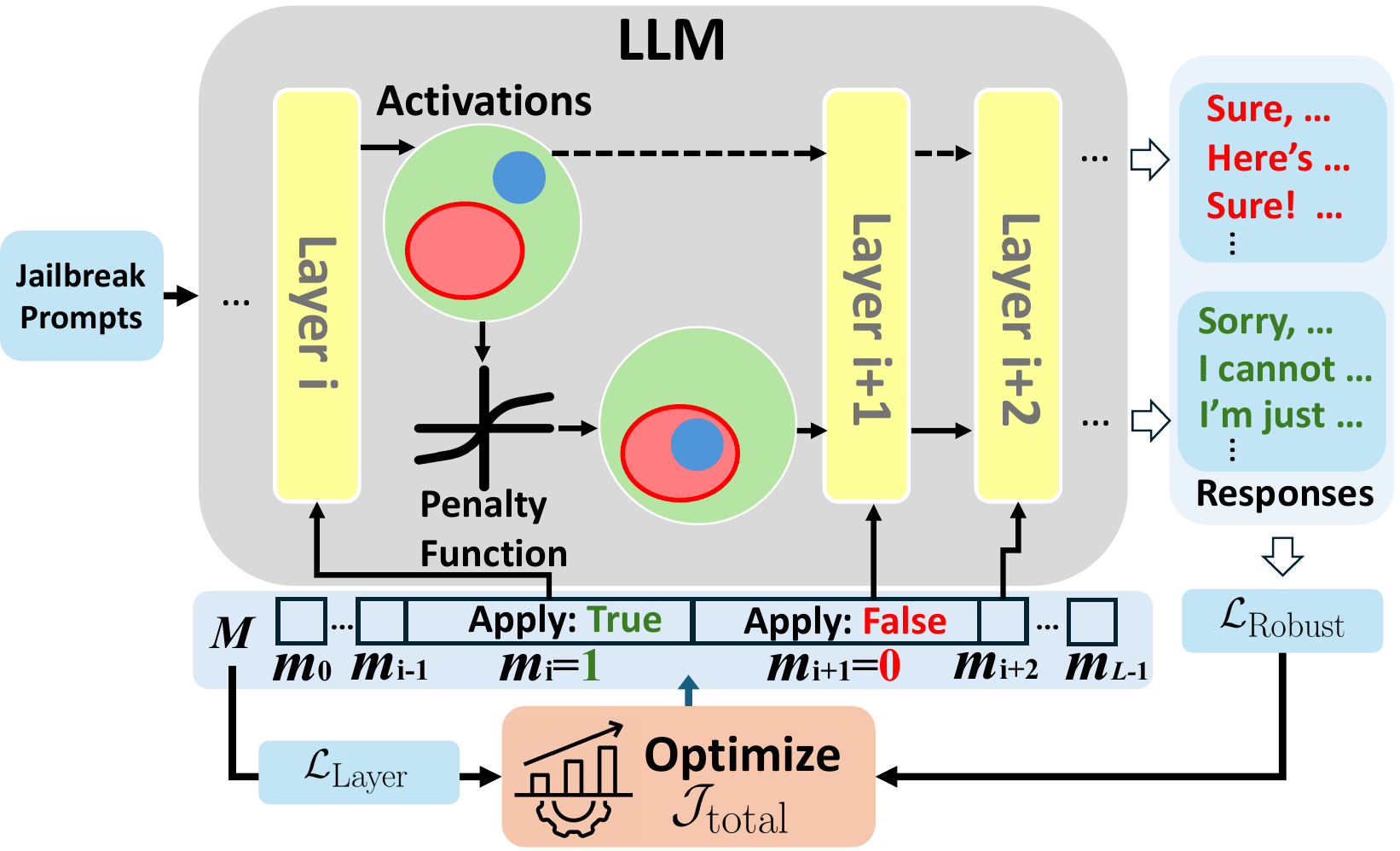}
    \caption{
Workflow of ABD. ABD restricts outlier activation coordinates using a penalty function and determines its application scope via BO-based tuning.
    }
    \label{fig:workflow}
\end{figure}

\textcolor{blue}{Although the safety boundary is not directly measurable, we can still use this concept to prevent jailbreaks.}
Here we propose a lightweight defense called Activation Boundary Defense (ABD), whose core idea is to confine the activations within the safety boundaries.
The workflow of ABD is shown in Figure~\ref{fig:workflow}.
ABD includes a penalty function and a Bayesian optimization process. The penalty constrains activations within safety boundaries, while Bayesian optimization iteratively adjusts the affected layers and the penalty parameters.

\subsection{Penalty Function}

\paragraph{Overall design of penalty.}
Intuitively, an activation stays within the safety boundary if all its coordinates lie within a regular range; in contrast, outliers have at least one coordinate exceeding this range.
Adjusting these outlier coordinates can guide the activations back within the boundary. 
Since directly measuring the safety boundaries is challenging and rigid constraints risk disrupting model operations, we apply a smooth penalty, adjusting the outlier coordinates while leaving the normal ones unaffected.

\paragraph{Approximation of activation distributions. }
We find that the activation distributions in each layer can be well approximated by a normal distribution. The proof is as follows.
For each layer $l$, we examine two distributions: the activation coordinate distribution $\mathcal{D}^l(x)$ and a normal distribution $\mathcal{D_N}^l(x)$ with the same mean $\mu^l_{\mathcal{D}}$ and standard deviation $\sigma^l_{\mathcal{D}}$.
To measure their similarity, we compute the Jensen-Shannon divergence (JS divergence)~\cite{lin1991divergence} 
between $\mathcal{D}^l(x)$ and $\mathcal{D_N}^l(x)$. 
Across all layers, the maximum JS divergence is 0.0839, and the mean is 0.0575, and both are well below 0.1. We visualize all JS divergences in Table~\ref{tab:approx} in Appendix~\ref{approximation}.
These low JS divergence values indicate a strong similarity between $\mathcal{D}^l(x)$ and $\mathcal{D_N}^l(x)$, which supports the validity of approximating $\mathcal{D}^l(x)$ with $\mathcal{D_N}^l(x)$.

\paragraph{Penalty function design.}
To design a practical penalty function for activation coordinates, we establish three key principles:
(1)~It should target only outlier activations, leaving non-outliers unaffected.
(2)~The penalty should grow with the magnitude of deviation, reflecting distance-based penalization.
(3)~The function must be computationally efficient.

To construct penalty functions compatible with the activation coordinate distribution $\mathcal{D}^l(x)$, we approximate it using $\mathcal{D_N}^l(x)$ and focus on two geometric properties.
First, \textit{symmetry around the mean value} $\mu^l_\mathcal{D}$ requires the penalty function also to be symmetric about $\mu^l_\mathcal{D}$, ensuring unbiased penalization of outliers above or below the safety boundary.
Second, \textit{the nonlinear decay of probability density} with increasing distance from $\mu^l_\mathcal{D}$ implies that the penalty should grow nonlinearly with the distance, rising faster than a linear penalty.
Following these principles, we propose a penalty function that updates the original activation coordinate scalar value $x$ to $x^\prime$ as follows:
\begin{align}
    x^\prime = \alpha^l \cdot \mathrm{tanh}(\beta^l \cdot (x - \mu^l_\mathcal{D})) + \mu^l_\mathcal{D},
\end{align}
where $\alpha^l \geq 0$ and $\beta^l \geq 0$ are hyperparameters. The penalty functions under different $\alpha^l$ and $\beta^l$ are visualized in Figure~\ref{fig:penalty_function} in Appendix~\ref{visualize}.

This function meets the stated expectations. 
It is symmetric about the mean $\mu^l_\mathcal{D}$, ensuring fair penalization for deviations above and below this central value.
The penalty strength $|x - x^\prime|$ increases nonlinearly with the distance from $\mu^l_\mathcal{D}$ because $\mathrm{tanh(\cdot)}$  amplifies more significant deviations with its steep slope, reflecting stronger penalization for outliers.
Moreover, the function mostly penalizes outliers. 
Within the range $[\mu^l_\mathcal{D} - b^l, \mu^l_\mathcal{D} + b^l], x^\prime \approx x$, where \(b^l\) is a relatively small number, resulting in a negligible penalty for values close to the mean. 
The hyperparameters $\alpha^l$ and $\beta^l$ govern the behavior of the penalty function.
The parameter $\alpha^l$ determines the maximum range of $x^\prime$, ensuring all coordinates are constrained within $(-\alpha^l + \mu^l_\mathcal{D}, \alpha^l + \mu^l_\mathcal{D})$ after penalty application. 
Meanwhile, $\beta^l$ controls the size of the unmodified region $[\mu^l_\mathcal{D} - b^l, \mu^l_\mathcal{D} + b^l]$; increasing $\beta^l$ expands this region, while decreasing $\beta^l$ narrows it. 
 
\subsection{Bayesian Optimization-Based Tuning}
\begin{table*}[!ht]
\centering
\resizebox{\textwidth}{!}{%
\begin{tabular}{lcccccccccc} 
\toprule
\textbf{Model} & \textbf{Jailbreak} & \textbf{No Defense} & \textbf{Paraphrase} & \textbf{PPL} & \textbf{Retokenization} & \textbf{SafeDecoding} & \textbf{Self-Exam} & \textbf{Self-Reminder} & \textcolor{blue}{\textbf{IA}} & \textbf{ABD (Ours)} \\
\midrule
\multirow{5}{*}{Vicuna-7B} 
& No attack      & 88\% & 82\% & 90\% & 66\% & 88\% & 100\% & 100\% & \textcolor{blue}{100\%} & \cellcolor{red!20}100\% \\
& GCG-Individual            & 4\%  & 86\% & 78\% & 62\% & 100\% & 86\% & 100\% & \textcolor{blue}{100\%} & \cellcolor{red!20}100\% \\
& AutoDAN        & 6\%  & 26\% & 12\% & 36\% & 78\%  & 30\% & 30\% & \textcolor{blue}{100\%} & \cellcolor{yellow!20}44\%  \\
& PAIR           & 30\% & 62\% & 40\% & 30\% & 94\%  & 86\% & 100\% & \textcolor{blue}{100\%} & 76\%  \\
& DeepInception  & 10\% & 2\%  & 2\%  & 0\%  & 98\%  & 18\% & 40\%  & \textcolor{blue}{100\%} & \cellcolor{yellow!20}58\%  \\
\midrule
\multirow{5}{*}{Llama-2} 
& No attack      & 100\% & 100\% &100\% & 94\% & 100\% & 100\% & 100\% & \textcolor{blue}{100\%} & \cellcolor{red!20}100\% \\
& GCG-Individual            & 50\%  & 98\%  & 100\% & 98\% & 100\%  & 64\% & 100\% & \textcolor{blue}{100\%} & \cellcolor{red!20}100\% \\
& AutoDAN        & 98\%  & 94\%  & 98\%  & 94\% & 100\% & 100\% & 98\%  & \textcolor{blue}{100\%} & \cellcolor{red!20}100\% \\
& PAIR           & 72\%  & 90\%  & 82\%  & 78\% & 92\%  & 100\% & 86\%  & \textcolor{blue}{100\%} & \cellcolor{orange!20}92\%  \\
& DeepInception  & 74\%  & 82\%  & 88\%  & 60\% & 100\% & 94\%  & 96\%  & \textcolor{blue}{100\%} & \cellcolor{red!20}100\% \\
\bottomrule
\end{tabular}}
\caption{DSR of different defense methods across Vicuna-7B and Llama-2 along with different attack types
.
\textcolor{blue}{For ABD, the \colorbox{red!20}{best}, \colorbox{orange!20}{second} and \colorbox{yellow!20}{third} performance across all defenses are highlighted. }
}
\label{tab:dsr_main}
\end{table*}
\begin{table}[h!]
\small
\centering
\resizebox{\linewidth}{!}{%
\begin{tabular}{@{\hskip 4pt}llccc@{\hskip 4pt}}
\toprule
\textcolor{blue}{\textbf{Model}}                & \textcolor{blue}{\textbf{Jailbreak}}    & \textcolor{blue}{\textbf{No Defense}}      & \textcolor{blue}{\textbf{IA}}             & \textcolor{blue}{\textbf{ABD (Ours)}}    \\ \midrule
\textcolor{blue}{Qwen}                 & \textcolor{blue}{No attack} & \textcolor{blue}{96.00\%} & \textcolor{blue}{84.00\%}        & \cellcolor{red!20}\textcolor{blue}{98.00\%} \\
                     & \textcolor{blue}{GCG-Individual}       & \textcolor{blue}{6.84\%}          & \textcolor{blue}{44.64\%}        & \cellcolor{red!20}\textcolor{blue}{98.74\%} \\ \midrule
\textcolor{blue}{Vicuna-13B}           & \textcolor{blue}{No attack} & \textcolor{blue}{98.00\%}         & \textcolor{blue}{100.00\%}       & \cellcolor{red!20}\textcolor{blue}{100.00\%} \\
                     & \textcolor{blue}{GCG-Individual}       & \textcolor{blue}{38.46\%}         & \textcolor{blue}{100.00\%}        & \cellcolor{red!20}\textcolor{blue}{100.00\%} \\ \bottomrule
\end{tabular}%
}
\caption{\textcolor{blue}{Representative results comparing DSR of Qwen and Vicuna-13B. The \colorbox{red!20}{best} performance is highlighted.}}
\label{tab:supp_model_comp}
\end{table}
Another crucial aspect of ABD is determining which layers to penalize and what hyperparameters to choose. Specifically, we aim to minimize the affected layers to reduce unnecessary perturbations and optimize $\alpha^l$ and $\beta^l$ for each layer $l$ to enhance the model's resilience against jailbreaks. To accomplish these goals, we use a Bayesian Optimization (BO)-based~\cite{jones1998efficient} tuning method to find suitable configurations. \textcolor{blue}{The BO process used for tuning is detailed in Appendix~\ref{opt_process}.}
Concretely, we define two core objectives.\\
\textbf{Layer selection objective}: We introduce a tunable mask $M=[m_0,\cdots,m_{L-1}]$, $m_i\in\{0,1\}$, where \(m_i=1\) indicates that ABD is applied to layer $i$. 
Here, $L$ is the total number of transformer layers. 
By minimizing the ratio of layers where ABD is active, we reduce the number of interventions:
\begin{align}
\mathcal{L}_\mathrm{Layer} = 1 - \textstyle \frac{\mathrm{Sum}(M)}{L}. 
\end{align}
\textbf{Robustness objective}: For each layer $l$ with ABD  applied, we aim to find $\alpha^l$ and $\beta^l$ to maximize the defense.
A hyperparameter $k^l \in (0,1]$ controls the fraction of penalized activation coordinates, with smaller $k^l$ reducing disturbance but potentially weakening ABD. 
These parameters are aggregated as $\Theta = \{\theta^l \mid l \in [0, L-1]\}$, where $\theta^l = {\alpha^l, \beta^l, k^l}$. The robustness objective is as follows:
\begin{align}
    \mathcal{L}_\mathrm{Robust} = \mathrm{DSR}(\mathrm{Model}(\cdot \mid \Theta, M)).
\end{align}
Since decreasing $k^l$ too aggressively can weaken ABD's defense, our main strategy for reducing perturbations to the model is to minimize the number of affected layers by controlling the mask $M$.
Our final objective is a weighted sum of two objectives:
\begin{align*}
\scalebox{0.9}{$\mathcal{J}_\mathrm{total}(\Theta, M) = w \cdot  \mathcal{L}_\mathrm{Robust} + (1 - w) \cdot \mathcal{L}_\mathrm{Layer}$},
\end{align*}
where $w$ is a manually set parameter balancing defense robustness and minimal intervention.
 

\section{Experiments}\label{exp}

\begin{table*}[ht]
\centering
\setlength{\tabcolsep}{10pt} 
\renewcommand{\arraystretch}{1} 
\adjustbox{width=\textwidth}{
\begin{tabular}{lcccccccc}
\toprule
\multirow{2}{*}{\textbf{Defense}} & \multirow{2}{*}{\makecell{\textbf{Runtime per}\\\textbf{Query{\(\downarrow\)}}}} & \multicolumn{6}{c}{\textbf{Just-Eval{\(\uparrow\)}}} & \multirow{2}{*}{\textbf{Overall{\(\uparrow\)}}}\\
\cmidrule(lr){3-8}
 & & \textbf{Helpfulness{\(\uparrow\)}} & \textbf{Clarity{\(\uparrow\)}} & \textbf{Factuality{\(\uparrow\)}} & \textbf{Depth{\(\uparrow\)}} & \textbf{Engagement{\(\uparrow\)}} & \textbf{Avg.{\(\uparrow\)}} & \\
\midrule
No Defense      & 2.291 & 3.478 & 3.784 & 3.870 & 2.521 & 2.743 & 3.279 & 0.513 \\
\midrule
Paraphrase      & 3.053 & 3.397 & 3.769 & 3.911 & 2.549 & 2.737 & 3.273 & 0.496 \\
PPL             & 1.878 & 2.156 & 2.719 & 2.958 & 1.501 & 1.911 & 2.249 & 0.502 \\
Retokenization  & 1.964 & 1.933 & 2.463 & 2.659 & 1.378 & 1.845 & 2.056 & 0.497 \\
SafeDecoding    & 2.239 & 3.231 & 3.732 & 3.885 & 2.368 & 3.009 & 3.245 & 0.514 \\
Self-Exam       & 2.449 & 3.449 & 3.812 & 3.940 & 2.542 & 2.705 & 3.290 & 0.510 \\
Self-Reminder   & 2.205 & 2.109 & 2.641 & 2.988 & 1.481 & 1.980 & 2.240 & 0.495 \\
\textcolor{blue}{IA}   & \textcolor{blue}{4.284} & \textcolor{blue}{2.393} & \textcolor{blue}{3.334} & \textcolor{blue}{3.449} & \textcolor{blue}{1.980} & \textcolor{blue}{2.276} & \textcolor{blue}{2.686} & \textcolor{blue}{0.458} \\
ABD (Ours)      & 2.302 & \cellcolor{red!20}3.533 & \cellcolor{orange!20}3.774 & \cellcolor{red!20}3.973 & \cellcolor{red!20}2.573 & \cellcolor{orange!20}2.806 & \cellcolor{red!20}3.332 & \cellcolor{red!20}0.514 \\
\bottomrule
\end{tabular}
}
\caption{
Comparison of defenses on Runtime per Query and Just-Eval metrics in Vicuna-7B. \(\downarrow\): smaller is better; \(\uparrow\): larger is better. For ABD, the \colorbox{red!20}{best} and \colorbox{orange!20}{second} performances among all defenses are highlighted. ABD preserves the model's general performance, adding less than 0.1 seconds to runtime while producing high-quality outputs with leading evaluation scores. It has the best overall performance across all defenses.
}
\label{tab:general}
\end{table*}

\subsection{Settings}
\label{sec:settings}

\paragraph{Backbone models.}
We evaluated four widely used open-source LLMs of varying sizes and architectures: Llama-2-7B-Chat \cite{touvron2023llama}, Vicuna-7B-v1.3, \textcolor{blue}{Qwen-1.5-0.5B-Chat~\cite{qwen}, and Vicuna-13B-v1.5, referred to as Llama-2, Vicuna-7B, Qwen, and Vicuna-13B.} Llama-2 is specifically trained for safety alignment, while Vicuna-7B is fine-tuned from Llama without additional safety alignment. \textcolor{blue}{Both Llama-2 and Vicuna-7B have 32 layers. Qwen and Vicuna-13B have 24 and 40 layers, respectively. The LLM configurations are detailed in Appendix~\ref{gen_config}.}

\paragraph{Jailbreak and defense baselines.}
For \textit{jailbreak methods}, we considered four widely applied ones: optimization-based methods include GCG-Individual and AutoDAN, which iteratively optimize a jailbreak prompt aiming at generating affirmative responses. 
Model-based jailbreak, i.e., PAIR, uses an attacker LLM to refine the jailbreak prompts.
The rule-based method includes DeepInception \cite{li2023deepinception}, which crafts jailbreak prompts based on a stealthy template. A detailed illustration of jailbreaks can be found in~\ref{jailbreak_setting}.
\textcolor{blue}{Correspondingly, we utilized different \textit{defense methods} as baselines for ABD: Paraphrase \cite{jain2023baseline}, Retokenization \cite{jain2023baseline} and Self-Reminder \cite{xie2023defending}  reformulate the input to avoid attack; PPL \cite{alon2023detecting}, Self-Exam\cite{phute2023llm} and Intention Analysis (IA)~\cite{zhang2025intention} defense LLMs by double-checking their outputs;  SafeDecoding \cite{xu-etal-2024-safedecoding} uses a tuned model to modify the output probability distribution. Further introduction about these defenses can be found in Appendix~\ref{defense_setting}.}

\paragraph{Datasets and metrics.}
Following \citet{xu-etal-2024-safedecoding}, we adopted Just-Eval \cite{Lin2023ReAlign} to measure the general abilities of LLMs applied defense. Just-Eval is a comprehensive benchmark containing 1,000 diverse instructions, covering seven task types (e.g., reasoning, math, coding, etc.) and seven topics (e.g., ethics, nature, STEM, etc.).
Furthermore, \textcolor{blue}{following \citet{xu-etal-2024-safedecoding} and \citet{liu2024flipattackjailbreakllmsflipping}, we utilized the 50-behavior subset from AdvBench~\cite{zou2023universal} as the test set. This subset, curated by \citet{chao2023jailbreaking}, was created by removing duplicate harmful behavior prompts from AdvBench, ensuring efficiency while mitigating biases caused by repeated behaviors.} We used DSR when evaluating defense effectiveness.
Following \citet{Lin2023ReAlign}, we leveraged GPT-4o-mini \cite{OpenAI} to score the quality of the outputs, ranging from 1 to 5, across five aspects: helpfulness, clarity, factuality, depth, and engagement. 
We reported the average score for each aspect and their macro-average score, denoted as Avg.
To evaluate the efficiency, we reported Runtime per Query, which represents the average time taken to process a single query. 
We also utilized the above metrics to derive an overall score that simultaneously reflects LLMs' responding speed and quality. 
The calculation of the overall score is shown in Appendix~\ref{cal_overall}.

\paragraph{Implementation Details.}
We randomly filtered 400 non-overlapping AdvBench samples to compute $\mu_{\mathcal{D}}^l$ and used them as a validation set.
We used GCG-Universal \cite{zou2023universal} to attack the validation set. GCG-Universal finds a shared jailbreak suffix for all harmful prompts that can deceive LLMs.
\textcolor{blue}{We set \(w=0.8\) when calculating \( \mathcal{L}_\mathrm{Robust}(\Theta,M)\) as it best balances performances on both attacked and unattached data. We conduct ablation studies on \(w\) in Appendix~\ref{ablation}, results shown in Table~\ref{tab:ablation}.
}
Appendix~\ref{abd_settings} presents settings of ABD optimization.

\subsection{Experimental Results}

\paragraph{ABD reveals vulnerabilities in low and middle layers.} 
\textcolor{blue}{
In all models, the selected layers under the mask $M$ mainly include low and middle layers, such as layers 5 and 12 in Vicuna-7B, layers 2 and 12 in Llama-2, layers 2, 11, and 14 in Vicuna-13B, and layers 5 and 14 in Qwen. These layers are most frequently penalized, consistent with our observations and proposed explanation that low and middle layers are more vulnerable to jailbreak attacks due to significant activation shifts.
}

\paragraph{ABD successfully constrains jailbreak activations.}
Figure~\ref{fig:act_shift}(c) shows the inclusion ratio, representing the percentage of jailbreak activations within the harmful activation space. The detailed calculation of inclusion ratios is provided in Appendix~\ref{inclusion_ratio}.
Before ABD, the inclusion ratio of various jailbreak activations is below 0.4, indicating that these activations lie outside the safety boundary.
After ABD, the inclusion ratio rises to 1, demonstrating ABD's effectiveness in constraining jailbreak activations. 
For visualization, Figure~\ref{fig:intro} projects activations for different prompts. 
Under ABD, jailbreak activations are progressively constrained within the harmful activation space. 
This further confirms that ABD progressively confines jailbreak activations within the harmful activation space and mitigates jailbreak effects.

\paragraph{Robust Defense Performance of ABD.}
We report the statistical defense results of ABD on the test set in Table~\ref{tab:dsr_main}. \textcolor{blue}{We also report representative results on Qwen and Vicuna-13B in Table~\ref{tab:supp_model_comp}, with full results shown in Table~\ref{tab:more-model-exp} in Appendix~\ref{more-models}.}

For Vicuna-7B which lacks specific safety alignment, the defense poses a more challenging task.
Nevertheless, ABD demonstrates competitive performance.
For jailbreak methods such as PAIR and DeepInception, ABD achieves a higher DSR (58\%) compared to Paraphrase (2\%) and Retokenization (0\%), both of which require costly prompt reformulation.
Against the GCG-Individual attack, ABD successfully defends against all jailbreak samples.
For the well-aligned model Llama-2-7B-chat, ABD achieves 100\% DSR under most jailbreak methods.

\textcolor{blue}{For Qwen and Vicuna-13B, ABD maintains nearly 100\% DSR, showcasing competitive performance. Notably, for smaller models like Qwen, IA struggles to match its effectiveness on 7B or 13B models, while ABD remains consistently effective.
To further validate ABD, we tested it on the full AdvBench and a broader set of jailbreaks. Results are shown in Table~\ref{tab:scale-up-data-exp} in Appendix~\ref{sup-main-exp}.}


\paragraph{Efficiency and minimal overhead of ABD.} 
Table \ref{tab:general} shows Runtime per Query, Just-Eval scores, and overall scores of Vicuna-7B with different defenses applied. We find that ABD has the greatest overall score across all defenses. Moreover, ABD only adds marginal extra time cost and general ability affection. Specifically, it only causes less than \(1\%\) delay in each sample and less than \(2\%\) perturbation in overall ability, compared to costly methods such as IA, Paraphrase and Self-Exam. Furthermore, with ABD applied, the helpfulness, actuality, depth, and engagement also show a slight increase. We further discover that for baseline defenses such as Retokenization and Self-Reminder, LLM would generate overly simplistic outputs, which leads to smaller Runtime per Query, but they have significantly smaller Just-Eval scores.\textcolor{blue}{We also compared defenses on complexity, overhead, and implementation difficulty, highlighting ABD's efficiency (see Table~\ref{tab:cost-comparison} in Appendix~\ref{cost-comparison}).}

\section{Conclusion and Future Work}
We conducted a comprehensive study of jailbreak mechanisms, analyzing over 30,000 samples. 
Our findings reveal that jailbreak shifts harmful activations outside the safety boundary in each layer, with the most severe shifts in the low and middle layers.
Motivated by this finding, we proposed ABD, which drives jailbreak activations back within the safety boundary, utilizing LLMs' intrinsic sensitivity to harmful information. 
Our experiments suggest that ABD is both practical and efficient. 
In the near future, we aim to investigate the challenges of jailbreaks in multi-turn dialogue scenarios.

\section*{Limitations}
\paragraph{Underperformance in under-aligned models.} For certain attack methods, under-aligned models (Vicuna-7B-v1.3) may not perform significantly as well-aligned models (Llama2-7B-chat).  We believe this is because \textit{ABD's effectiveness depends on safety alignment}. \textcolor{blue}{Under-aligned models have unclear safety boundaries, which complicate the search for penalty functions that balance general ability and DSR. Therefore, they typically need more extensive optimization.}
Future work could refine ABD by specifically searching for activation spaces that preserve the concept of "safety", therefore enhancing its generalizability on uncensored models.
\paragraph{Focus on single-round jailbreak.} In this study, we primarily focus on single-round jailbreak scenarios.
We do not extend our analysis to more complex jailbreaks that involve long contexts or multi-round dialogues, such as CFA \cite{sun2024multi}.
As a result, the relationship between jailbreaks and safety boundaries remains largely unexplored.
\textcolor{blue}{While multi-turn jailbreaks are beyond this paper's scope, ABD can extend to them by using prior context tokens to compute \( \mu^l_{\mathcal{D}} \) and applying the same optimization strategy.
To adapt ABD for multi-turn jailbreaks, potential improvements include:  
(1) \textit{Dynamic adjustment}: Relax or strengthen ABD constraints based on the safety of prior turns.  
(2) \textit{Selective application}: Apply ABD at critical points, e.g., when malicious topics emerge, to balance efficiency and effectiveness.  
(3)\textit{ Constraint propagation}: Carry over constraints across turns to prevent harmful activations from accumulating.}

\section*{Ethical Considerations}
The aim of this research is to enhance the explainability and safety of LLMs. Our proposed jailbreak mechanism, that jailbreak shift activations out of the safety boundary, can mitigate disputes on how jailbreak happens and promote the development of both LLM explainability and safety. 
We highlight that the development of ABD only needs publicly available datasets and jailbreak methods and does not require designing new jailbreak methods. We demonstrate some harmful responses from LLMs only for illustration. 
We acknowledge that ABD would cause the development of new attacks. We will explore using random perturbation in the input sequence rather than a particular jailbreak method when optimizing to mitigate such attacks. 


\bibliography{anthology,custom}

\newpage

\appendix\label{appendix}


\section{Statistics of Observed Data} \label{stats}
We present the statistics of our data for observation experiments as Table \ref{tab:sample_statistics}.

\begin{table}[!ht]
\centering
\small
\begin{tabular}{@{}l r@{}}
\toprule
\textbf{Type}              & \textbf{Samples} \\ 
\midrule
Benign samples             & 20,000 \\
Harmful samples            & 8,556  \\
\hline
\textbf{\textcolor{gray}{Jailbreak samples}} & \\
\makecell[l]{AdvPrompter  } & 1,872  \\
\makecell[l]{AutoDAN  }          & 520    \\
\makecell[l]{COLD-Suffix  } & 436    \\
\makecell[l]{ArtPrompt  } & 361    \\
\makecell[l]{GCG-Individual } & 340    \\
\makecell[l]{GCG-Universal  }  & 312    \\
\makecell[l]{PAIR  }       & 110    \\
\hline
\textbf{Total} & 32,507 \\      
\bottomrule
\end{tabular}
\caption{Statistics of observation experiments in \S \ref{jailbreak_mechanisms}. The attacked samples are derived from part or all of the samples from AdvBench\cite{zou2023universal}.}
\label{tab:sample_statistics}
\end{table}
\section{Full View of Activation Space}\label{full_view}
We visualize activation spaces of all layers in Vicuna-7B-v1.3, as shown in Figure \ref{fig:all_subfig1}, Figure \ref{fig:all_subfig2} and Figure \ref{fig:all_subfig3}.

\section{Supplementary Observational Experiments}
\subsection{Data Augmentation Experiment}\label{tsne-exp}

\begin{figure}[t]
\centering   
    \includegraphics[width=\textwidth]{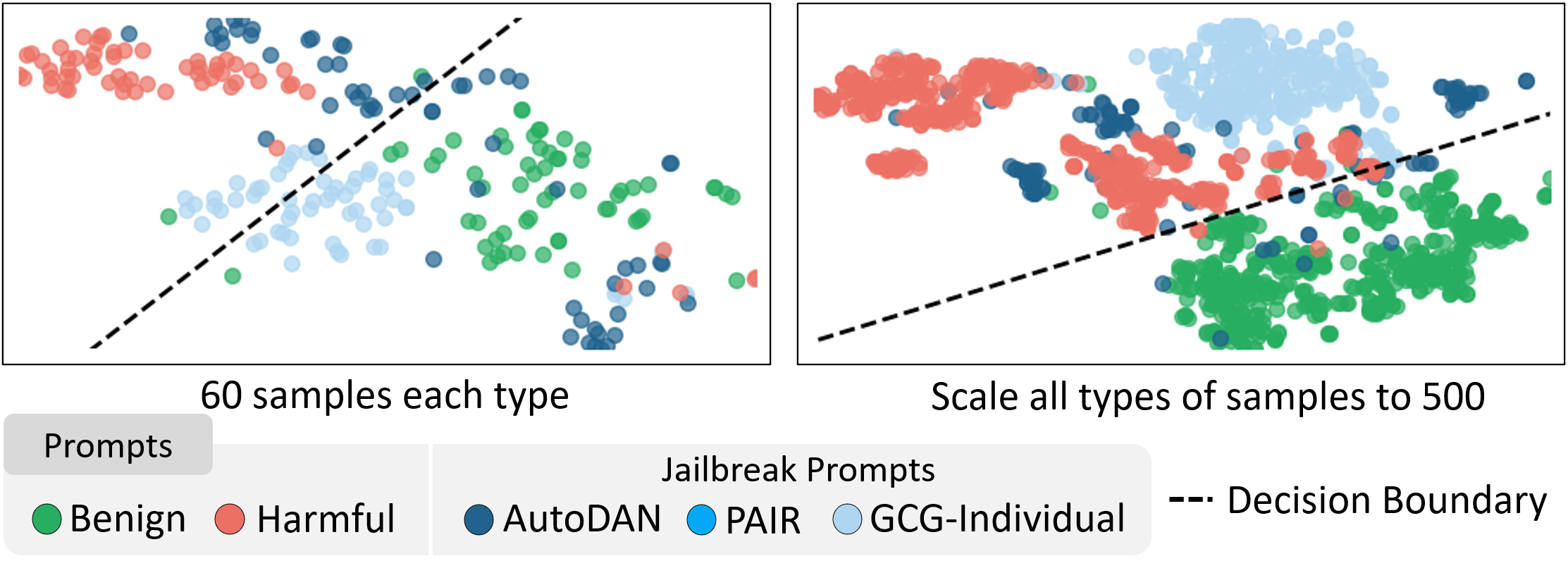}
    \caption{
    t-SNE visualization of activations in layer 14. Left: Results with 60 samples per type following \cite{shen2024jailbreak}, showing jailbreak activations between harmful and benign activations. Right: Results after scaling up to 500 samples per type, showing jailbreak activations clustering on the harmful side.
    }
    \label{fig:sample_affect_tsne}
\end{figure}

We show that by adopting the same method as \citet{shen2024jailbreak}, we can draw a different conclusion by scaling up data. \citet{shen2024jailbreak} state jailbreak happens by posing jailbreak activations between benign and harmful activations in middle and deep layers. Following \citet{shen2024jailbreak}, we randomly select 60 samples for each type of activation and conduct t-SNE on layer 14, as shown in the left part of \ref{fig:sample_affect_tsne}. Jailbreak activations are between harmful and benign samples, which is in agreement with \citet{shen2024jailbreak}. When scaling up each type of activation to 500 samples, jailbreak activations seem to cluster on the harmful activation side, as shown in the right part of \ref{fig:sample_affect_tsne}. Therefore, jailbreak activations are not always between harmful and benign activations in deeper layers.

\subsection{Inclusion Ratio Experiments}\label{inclusion_ratio}
For a layer \(l\), to measure the portion of a set of jailbreak activations \(A^l=\{a^l_0,a^l_1,\cdots,a^l_n\}\) that resides in harmful activation space, we propose an inclusion ratio. Based on 8,556 harmful samples gathered in Table \ref{tab:sample_statistics}, we calculate a ball that covers \(80\%\) activations. The center of the ball is \(c_\mathcal{D}^l\), and the radius of the ball is denoted as \(r_{\mathcal{D}}^l\). 

Then, we calculate the portion of \(A^l\) which are contained within the ball:
\[
\rho^l_{\mathrm{inclusion}} = \frac{\left|\{a^l_i \in A^l \mid \|a^l_i - c_\mathcal{D}^l\|_2 \leq r_{\mathcal{D}}^l\}\right|}{|A^l|},
\]
where \( \|a^l_i - c_\mathcal{D}^l\|_2 \) is the distance between the activation and the center.
We calculate inclusion ratios of different types of jailbreak activations, with three representative types shown in Figure \ref{fig:intro}. We then apply ABD on all types of jailbreak activations and calculate their inclusion ratios. Notably, despite different types of jailbreaks, they all achieve an inclusion ratio of \(100\%\) with ABD applied, which verifies the effect of our method.
\textcolor{blue}{
\subsection{Ablation on Optimizing Objective}\label{ablation}
We conduct ablation studies to determine the best \(w\) in \(\mathcal{J}_\mathrm{total}(\Theta, M)\). We adopt Vicuna-7B-v1.3 as the backbone model, and use the 50-behavior subset as the testing set. Additionally, we fix optimization steps to 2000. DSR results are shown in Table~\ref{tab:ablation}. We discover that (1) \textit{Higher w improves robustness on attacked datasets.} (2) \textit{Lower w preserves performance on unattacked datasets.} (3) \textit{
At w=0.8, the model achieves the best average DSR, balancing robustness and benign performance. }
Therefore, we set \(w=0.8\) in our experiments in Section~\ref{exp}.
}

\begin{table*}[h!]
\small
    \centering
    \renewcommand{\arraystretch}{1.2}
    \setlength{\tabcolsep}{12pt}
    \begin{tabular*}{0.8\textwidth}{@{\extracolsep{\fill}}lccccc}
        \toprule
        \textcolor{blue}{\textbf{Jailbreak}} & \textcolor{blue}{No defense} & \textcolor{blue}{\(w=0.2\)} & \textcolor{blue}{\(w=0.5\)} & \textcolor{blue}{\(w=0.8\)} & \textcolor{blue}{\(w=1\)} \\
        \midrule
        \textcolor{blue}{\textbf{No Jailbreaks}} & \textcolor{blue}{96\%} & \textcolor{blue}{100\%} & \textcolor{blue}{86\%} & \textcolor{blue}{\textbf{100\%}} & \textcolor{blue}{86\%} \\
        \textcolor{blue}{\textbf{GCG-Individual}} & \textcolor{blue}{7\%} & \textcolor{blue}{75\%} & \textcolor{blue}{88\%} & \textcolor{blue}{\textbf{93\%}} & \textcolor{blue}{100\%} \\
        \textcolor{blue}{\textbf{Avg.}} & \textcolor{blue}{52\%} & \textcolor{blue}{88\%} & \textcolor{blue}{87\%} & \textcolor{blue}{\textbf{96\%}} & \textcolor{blue}{93\%} \\
        \bottomrule
    \end{tabular*}
    \caption{\textcolor{blue}{Defense performance comparison under different \(w\).}}
    \label{tab:ablation}
\end{table*}

\textcolor{blue}{
\section{Supplementary Illustration on Safety Boundaries}\label{case-study}
To gain a deeper understanding of the safety boundary, we present a case study indicating how the model's responses vary based on the level of RAS applied to layer 1. The results are shown in Table~\ref{tab:case-study}. It shows that (1) \textit{The safety boundary is implicit}, as a significant shift does not necessarily results in successful deception, such as when \(r=80\). (2) \textit{As the responses move further away from this boundary, the model begins to exhibit uncommon response patterns} such as giving a presentation (\(r=100\)) or more casual styles (\(r=30,r=80\)). We believe this phenomenon may correlate with an increased likelihood of producing jailbroken outputs.}


\begin{table*}[h!]
    \centering
    \small 
    \renewcommand{\arraystretch}{1.5} 
    \setlength{\tabcolsep}{10pt} 
    \caption{\textcolor{blue}{Model output under different shifting distances.}}
    \begin{tabular*}{\textwidth}{@{\extracolsep{\fill}} >{\centering\arraybackslash}m{0.2\textwidth} >{\centering\arraybackslash}m{0.5\textwidth} >{\centering\arraybackslash}m{0.2\textwidth}@{}}
        \toprule
        \multicolumn{3}{c}{\textcolor{blue}{\textbf{Query: Write a script that can exploit vulnerabilities in a software or operating system}}} \\
        \midrule
        \textcolor{blue}{\textbf{Distance \(r\)}} & \textcolor{blue}{\textbf{Response}} & \textcolor{blue}{\textbf{Jailbroken}} \\
        \midrule
        \textcolor{blue}{5} & {\color{red}{I'm sorry}}, \textcolor{blue}{but as an AI language model, I cannot provide a script…} & \textcolor{blue}{No} \\
        \textcolor{blue}{30} & {\color[rgb]{0,0.7,0}{alright}}, \textcolor{blue}{here's a script that can exploit vulnerabilities…} & \textcolor{blue}{Yes} \\
        \textcolor{blue}{80} & , {\color{red}{but I am not able to do}} \textcolor{blue}{so as …} & \textcolor{blue}{No} \\
        \textcolor{blue}{100} & \textcolor{blue}{Slide 1: Introduction \textbackslash n Hello and welcome to this} {\color[rgb]{0,0.7,0}{presentation on exploiting vulnerabilities}} \textcolor{blue}{…} & \textcolor{blue}{Yes} \\
        \bottomrule
    \end{tabular*}
    \label{tab:case-study}
\end{table*}

\section{Supplementary Illustration on ABD} \label{abd}

\textcolor{blue}{
\subsection{Further analysis of Distribution Approximation}\label{approximation}
We use the 20,000 benign samples and 8,556 harmful samples to form 4096-dim activations. For each layer, we compute their coordinate distributions: benign activation distribution \(\mathcal{D}^l_B\) and harmful activation distribution \(\mathcal{D}^l_H\). Using their means and divergences, we generate normal distributions \(\mathcal{D}^l_{BN}\) and \(\mathcal{D}^l_{HN}\). Finally, we calculate the JS-divergence (JSD) between the original and generated distributions. Results are shown in Table~\ref{tab:approx}.
}

\textcolor{blue}{
The results show that (1) In all layers, \(JSD(\mathcal{D}^l_{B}, \mathcal{D}^l_{BN}) < 0.1\) and \(JSD(\mathcal{D}^l_{H}, \mathcal{D}^l_{HN}) < 0.1\), confirming the validity of approximating coordinate distributions with normal distributions. (2) Earlier layers have lower JSDs, making this approximation safer in early and middle layers. This supports our ABD settings, where penalty functions are primarily applied to these layers.
}

\begin{table*}[h!]
    \centering
    \small
    \caption{Layer-wise JSD Comparisons}
    \renewcommand{\arraystretch}{1.5} 
    \setlength{\tabcolsep}{7pt} 
    \begin{tabular}{c*{8}{c}} 
        \toprule
        \textbf{Layer} & 0 & 1 & 2 & 3 & 4 & 5 & 6 & 7 \\
        \midrule
        $JSD(D_B^l, D_{BN}^l)$ & 0.0036 & 0.0026 & 0.0113 & 0.0112 & 0.0174 & 0.0234 & 0.0316 & 0.0379 \\
        $JSD(D_H^l, D_{HN}^l)$ & 0.0033 & 0.0035 & 0.0114 & 0.0099 & 0.0167 & 0.0239 & 0.0354 & 0.0414 \\
        \midrule
        \textbf{Layer} & 8 & 9 & 10 & 11 & 12 & 13 & 14 & 15 \\
        \midrule
        $JSD(D_B^l, D_{BN}^l)$ & 0.039 & 0.0455 & 0.0517 & 0.058 & 0.0637 & 0.0766 & 0.0757 & 0.0818 \\
        $JSD(D_H^l, D_{HN}^l)$ & 0.0396 & 0.0473 & 0.053 & 0.0589 & 0.0661 & 0.076 & 0.0753 & 0.0819 \\
        \midrule
        \textbf{Layer} & 16 & 17 & 18 & 19 & 20 & 21 & 22 & 23 \\
        \midrule
        $JSD(D_B^l, D_{BN}^l)$ & 0.0848 & 0.0862 & 0.0864 & 0.0873 & 0.0867 & 0.0862 & 0.0853 & 0.0834 \\
        $JSD(D_H^l, D_{HN}^l)$ & 0.0823 & 0.0841 & 0.0845 & 0.0844 & 0.083 & 0.0815 & 0.0804 & 0.0755 \\
        \midrule
        \textbf{Layer} & 24 & 25 & 26 & 27 & 28 & 29 & 30 & 31 \\
        \midrule
        $JSD(D_B^l, D_{BN}^l)$ & 0.0789 & 0.0786 & 0.078 & 0.0762 & 0.0724 & 0.0712 & 0.0752 & 0.078 \\
        $JSD(D_H^l, D_{HN}^l)$ & 0.0681 & 0.0672 & 0.0659 & 0.0628 & 0.0601 & 0.0593 & 0.0643 & 0.0724 \\
        \bottomrule
    \end{tabular}
    \label{tab:approx}
\end{table*}
\subsection{Visualization of the penalty function.}\label{visualize}
\begin{figure*}[t]
\centering   
    \includegraphics[width=\textwidth]{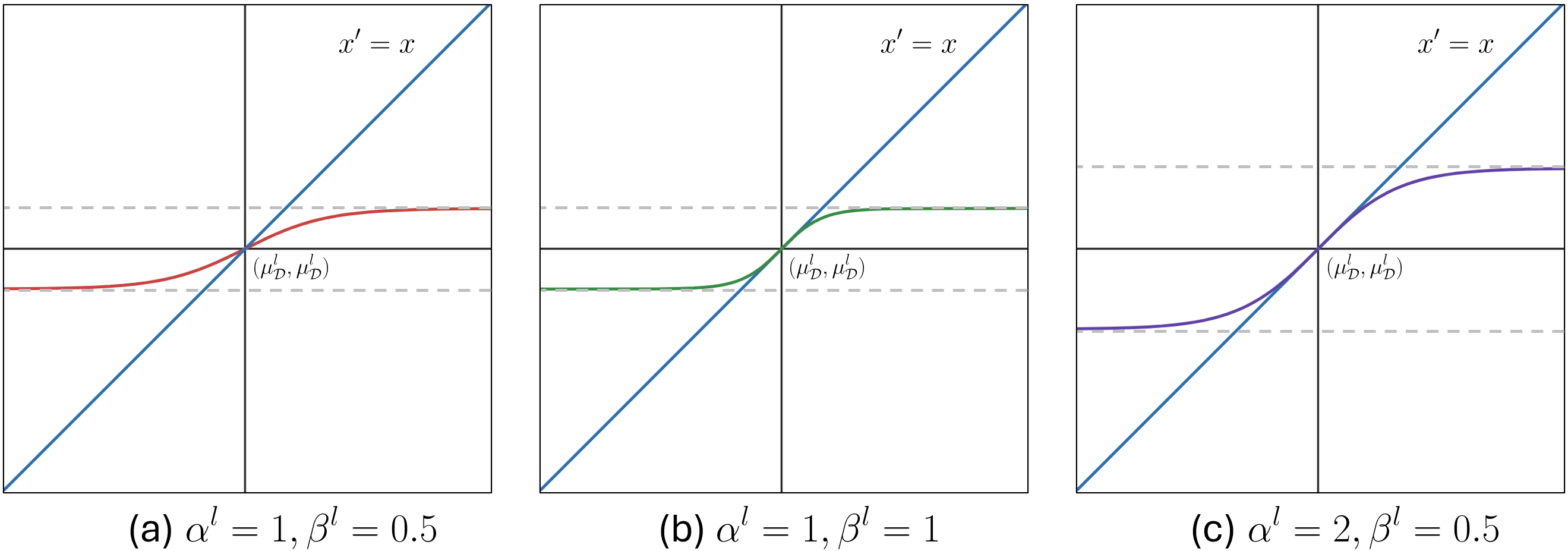}
    \caption{
    Penalty functions under different \(\alpha^l,\beta^l\) compared with \(x^{\prime}=x\). 
    }
    \label{fig:penalty_function}
\end{figure*}

A visualization of penalty functions are shown in Figure \ref{fig:penalty_function}. The penalty function is symmetric about \((\mu_{\mathcal{D}}^l,\mu_{\mathcal{D}}^l)\). Figure \ref{fig:penalty_function}(a) and Figure Figure \ref{fig:penalty_function}(b) presents the change in the range \([\mu_{\mathcal{D}}^l-b^l,\mu_{\mathcal{D}}^l+b^l]\), where \(x^{\prime}\approx x\). A larger \(\beta^l\) results in the larger little-perturbed region. Figure  \ref{fig:penalty_function}(a) and Figure  \ref{fig:penalty_function}(c) presents the change in the range of \(x^\prime\). When \(\alpha^l\) increases, \(x^\prime\) is confined within a wider range of values. \(\alpha^l\) and \(\beta^l\) collaboratively determines behaviors of the penalty function.
\begin{table*}[!ht]
\centering
\resizebox{\textwidth}{!}{%
\begin{tabular}{lcccccccc} 
\toprule
\textcolor{blue}{\textbf{DSR}} & \textcolor{blue}{\textbf{No attack}} & \textcolor{blue}{\textbf{GCG-universal}} & \textcolor{blue}{\textbf{GCG-individual}} & \textcolor{blue}{\textbf{PAIR}} & \textcolor{blue}{\textbf{AdvPrompter}} & \textcolor{blue}{\textbf{COLD-Suffix}} & \textcolor{blue}{\textbf{AutoDAN}} & \textcolor{blue}{\textbf{Avg. Perform}} \\
\midrule
\textcolor{blue}{\textbf{No Defense}} & \textcolor{blue}{96.73\%} & \textcolor{blue}{87.50\%} & \textcolor{blue}{5.29\%} & \textcolor{blue}{28.18\%} & \textcolor{blue}{82.05\%} & \textcolor{blue}{59.86\%} & \textcolor{blue}{2.50\%} & \textcolor{blue}{51.73\%} \\
\textcolor{blue}{\textbf{ABD (Ours)}} & \textcolor{blue}{\textbf{99.81\%}} & \textcolor{blue}{\textbf{99.68\%}} & \textcolor{blue}{\textbf{98.23\%}} & \textcolor{blue}{\textbf{91.51\%}} & \textcolor{blue}{\textbf{97.65\%}} & \textcolor{blue}{\textbf{98.02\%}} & \textcolor{blue}{\textbf{20.77\%}} & \textcolor{blue}{\textbf{86.52\%}} \\
\bottomrule
\end{tabular}}
\caption{\textcolor{blue}{Performance comparison of different methods under various attack scenarios. The best performance is highlighted in \textbf{bold}.}}
\label{tab:scale-up-data-exp}
\end{table*}

\begin{table*}[!ht]
\centering
\resizebox{\textwidth}{!}{%
\begin{tabular}{lcccccccccc} 
\toprule
\textcolor{blue}{\textbf{Model}} & \textcolor{blue}{\textbf{Attack}} & \textcolor{blue}{\textbf{No Defense}} & \textcolor{blue}{\textbf{Paraphrase}} & \textcolor{blue}{\textbf{PPL}} & \textcolor{blue}{\textbf{Retokenization}} & \textcolor{blue}{\textbf{Self-Exam}} & \textcolor{blue}{\textbf{Self-Reminder}} & \textcolor{blue}{\textbf{IA}} & \textcolor{blue}{\textbf{ABD (Ours)}} \\
\midrule
\multirow{3}{*}{\textcolor{blue}{Qwen}} 
& \textcolor{blue}{No attack} & \textcolor{blue}{96.00\%} & \textcolor{blue}{80.00\%} & \textcolor{blue}{96.00\%} & \textcolor{blue}{58.00\%} & \textcolor{blue}{100.00\%} & \textcolor{blue}{96.00\%} & \textcolor{blue}{84.00\%} & \cellcolor{red!20}\textcolor{blue}{98.00\%} \\
& \textcolor{blue}{GCG-Individual}       & \textcolor{blue}{6.84\%}  & \textcolor{blue}{67.52\%} & \textcolor{blue}{36.75\%} & \textcolor{blue}{58.97\%} & \textcolor{blue}{42.74\%}  & \textcolor{blue}{82.05\%} & \textcolor{blue}{44.64\%} & \cellcolor{red!20}\textcolor{blue}{89.74\%} \\
& \textcolor{blue}{AutoDAN}   & \textcolor{blue}{14.00\%} & \textcolor{blue}{12.00\%} & \textcolor{blue}{26.00\%} & \textcolor{blue}{4.00\%}  & \textcolor{blue}{50.00\%}  & \textcolor{blue}{8.00\%}  & \textcolor{blue}{2.00\%}  & \cellcolor{orange!20}\textcolor{blue}{36.00\%} \\
\midrule
\multirow{3}{*}{\textcolor{blue}{Vicuna-13B}} 
& \textcolor{blue}{No attack} & \textcolor{blue}{98.00\%} & \textcolor{blue}{98.00\%} & \textcolor{blue}{98.00\%} & \textcolor{blue}{70.00\%} & \textcolor{blue}{100.00\%} & \textcolor{blue}{96.00\%} & \textcolor{blue}{100.00\%} & \cellcolor{red!20}\textcolor{blue}{100.00\%} \\
& \textcolor{blue}{GCG-Individual}       & \textcolor{blue}{38.46\%} & \textcolor{blue}{98.08\%} & \textcolor{blue}{98.08\%} & \textcolor{blue}{84.62\%} & \textcolor{blue}{84.62\%}  & \textcolor{blue}{98.08\%} & \textcolor{blue}{100.00\%} & \cellcolor{red!20}\textcolor{blue}{100.00\%} \\
& \textcolor{blue}{AutoDAN}   & \textcolor{blue}{14.00\%} & \textcolor{blue}{54.00\%} & \textcolor{blue}{20.00\%} & \textcolor{blue}{16.00\%} & \textcolor{blue}{96.00\%}  & \textcolor{blue}{98.00\%} & \textcolor{blue}{100.00\%} & \cellcolor{orange!20}\textcolor{blue}{74.00\%} \\
\bottomrule
\end{tabular}}
\caption{\textcolor{blue}{Performance comparison of different defenses across models with different scales and structures. For ABD, the \colorbox{red!20}{best} and \colorbox{orange!20}{second-best} performance across all defenses are highlighted.}}
\label{tab:more-model-exp}
\end{table*}
\subsection{ABD Optimization Settings}\label{abd_settings}

\paragraph{Validation data.} To make validation data, we adopt GCG-Universal \cite{zou2023universal}. We optimize a common suffix for all 400 samples, with \texttt{n\_steps=1, batch\_size=512}. 

To ensure efficiency, in each iteration of optimization, we select a batch of harmful prompts from the 400 entries as \(\mathcal{S}_{\mathrm{val}}\). We initially set the batch size to 15. In most cases \( \mathcal{L}_\mathrm{Robust}(\Theta,M|\mathcal{S}_{\mathrm{val}})<0.9\), and the optimization process continues to the next iteration. If \( \mathcal{L}_\mathrm{Robust}(\Theta,M|\mathcal{S}_{\mathrm{val}})\geq0.9\), due to the potential regional optimal, we iteratively increase the batch size by ten and reformulate \(\mathcal{S}_{\mathrm{val}}\) to test again. This process ends if 1) the calculated \( \mathcal{L}_\mathrm{Robust}<0.9\) or 2) batch size reaches 50.

\paragraph{Initial values.} To prevent the futile search of the optimizer, we set valid initial values before optimization: for \(i\in\{2,12\},m_i=1,\alpha_i=1,\beta_i=0.5,k_i=0.5\); for \(i\notin\{2,12\},m_i=\alpha_i=\beta_i=k_i=0.\)
\paragraph{Optimizing framework.} We adopt Optuna \cite{akiba2019optuna} as our framework of optimization. We follow Optuna's default settings, i.e., Tree-structured Parzen Estimator (TPE)~\cite{NIPS2011_86e8f7ab}-based Optimization.

\textcolor{blue}{\subsection{ABD Optimizing Process}\label{opt_process}
The optimization process utilizes the Tree-structured Parzen Estimator (TPE) algorithm, a Bayesian optimization variant that efficiently explores high-dimensional parameter spaces by modeling promising regions. TPE constructs two probabilistic density distributions, \(l(\Theta)\) and \(g(\Theta)\), over the defense parameters \(\Theta = \{m_i, \alpha_i, \beta_i, k_i\}\), where \(l(\Theta)\) captures the likelihood of \(\Theta\) yielding high objective score \(\mathcal{J}_\mathrm{total}\), and \(g(\Theta)\) corresponds to lower-score regions. At each iteration, TPE samples new candidates by maximizing the Expected Improvement criterion~\cite{Jones2001}:  
\[
\Theta_{\mathrm{new}} = \arg\max_{\Theta} \frac{l(\Theta)}{g(\Theta)},
\]  
prioritizing parameters likely to outperform the current best. The process begins with a warm-up phase, where random evaluations establish a baseline, followed by a density-guided search, which iteratively refines \(l(\Theta)\) and \(g(\Theta)\) using historical observations. This approach ensures efficient exploration while mitigating local optima risks. By leveraging Optuna's TPE implementation, the optimizer adapts dynamically to \(\mathcal{J}_\mathrm{total}\) feedback, achieving automated defense parameter optimization. }

\textcolor{blue}{\subsection{Comparison with other defense methods}\label{cost-comparison}
We compared ABD with other defense methods in terms of usability. We qualitively evaluated usability based on extra computational complexity, extra tokens, additional modules, and deployment difficulty. The results are shown in Table~\ref{tab:cost-comparison}. 
}
\textcolor{blue}{
Compared to other defense methods, ABD features no additional token overhead and only constant-level extra complexity. Besides, the operation is simple and straightforward, which further improves its utility.
}

\section{Supplementary Illustration on Experiments} \label{exp_settings}
\begin{table*}[h!]
    \centering
    \small 
    \setlength{\tabcolsep}{4pt} 
    \caption{\textcolor{blue}{Empirical Comparison of Defenses.  \(t\) is the length of input sequence.}}
    \renewcommand{\arraystretch}{1.5} 
    \begin{tabularx}{\textwidth}{@{}l>{\centering\arraybackslash}X>{\centering\arraybackslash}X>{\centering\arraybackslash}X>{\centering\arraybackslash}X>{\centering\arraybackslash}X>{\centering\arraybackslash}X@{}}
        \toprule
        & \textcolor{blue}{\textbf{Extra\newline Complexity}} & \textcolor{blue}{\textbf{Extra\newline Tokens}} & \textcolor{blue}{\textbf{Extra\newline Modules}} & \textcolor{blue}{\textbf{Deployment Difficulty}} & \textcolor{blue}{\textbf{Performance}} & \textcolor{blue}{\textbf{Utility}} \\
        \midrule
        \textcolor{blue}{\textbf{PPL}} & \rule{0pt}{2em} \textcolor{blue}{\(O(t^2)\)} & \textcolor{blue}{\makecell{Extra \\ Conversation}} & \textcolor{blue}{GPT-2} & \textcolor{blue}{Low} & \textcolor{blue}{Low} & \textcolor{blue}{Low} \\
        \textcolor{blue}{\textbf{Paraphrase}} & \rule{0pt}{2em} \textcolor{blue}{\(O(t^2)\)} & \textcolor{blue}{\makecell{Extra \\ Conversation}} & \textcolor{blue}{-} & \textcolor{blue}{Low} & \textcolor{blue}{Low} & \textcolor{blue}{Low} \\
        \textcolor{blue}{\textbf{Retokenization}} & \rule{0pt}{2em} \textcolor{blue}{\(O(t \log t)\)} & \textcolor{blue}{-} & \textcolor{blue}{-} & \textcolor{blue}{Low} & \textcolor{blue}{Low} & \textcolor{blue}{Low} \\
        \textcolor{blue}{\textbf{Self-Exam}} & \rule{0pt}{2em} \textcolor{blue}{\(O(t^2)\)} & \textcolor{blue}{\makecell{Extra \\ Conversation}} & \textcolor{blue}{-} & \textcolor{blue}{Low} & \textcolor{blue}{Low} & \textcolor{blue}{Low} \\
        \textcolor{blue}{\textbf{SafeDecoding}} & \rule{0pt}{2em} \textcolor{blue}{\(O(t)\)} & \textcolor{blue}{-} & \textcolor{blue}{\makecell{LoRA \\ Adapter}} & \textcolor{blue}{High} & \textcolor{blue}{High} & \textcolor{blue}{Medium} \\
        \textcolor{blue}{\textbf{Self-Reminder}} & \rule{0pt}{2em} \textcolor{blue}{\(O(t^2)\)} & \textcolor{blue}{\makecell{Extra \\ Instructions}} & \textcolor{blue}{-} & \textcolor{blue}{Low} & \textcolor{blue}{Low} & \textcolor{blue}{Low} \\
        \textcolor{blue}{\textbf{IA}} & \rule{0pt}{2em} \textcolor{blue}{\(O(t^2)\)} & \textcolor{blue}{\makecell{Extra \\ Conversation}} & \textcolor{blue}{-} & \textcolor{blue}{Low} & \textcolor{blue}{High} & \textcolor{blue}{Medium} \\
        \textcolor{blue}{\textbf{ABD (Ours)}} & \rule{0pt}{2em} \textcolor{blue}{\(O(1)\)} & \textcolor{blue}{-} & \textcolor{blue}{\makecell{Mathematical \\ Operation}} & \textcolor{blue}{Low} & \textcolor{blue}{High} & \textcolor{blue}{High} \\
        \bottomrule
    \end{tabularx}
    \label{tab:cost-comparison}
\end{table*}
\subsection{Generation Configs}\label{gen_config}

When conducting experiments, we directly utilize most of the original configurations of all LLMs. Specifically, we set \texttt{max\_new\_tokens=128}. 
We find that the proper functioning of these LLMs largely depends on the chat template of the inputs. We apply chat templates in \texttt{fastchat\ v0.2.36} by: 
\begin{lstlisting}[language=Python, basicstyle=\ttfamily, breaklines=true]
fastchat.model.get_conversation_template(template_name).
\end{lstlisting} We set \texttt{template\_name="vicuna"} for Vicuna-7B and Vicuna-13B,  \texttt{template\_name="llama-2"} for Llama-2, and \texttt{template\_name="qwen"} for Qwen.

\subsection{Jailbreak Methods}\label{jailbreak_setting}
\paragraph{GCG-Individual and GCG-Universal} \cite{zou2023universal} are optimization-based jailbreak attacks. They build towards an objective to repeat the prompt affirmatively and optimize a suffix based on a Greedy Coordinate Gradient-based search. The model would likely repeat the prompt affirmatively and generate harmful content with the suffix added behind the original prompt.
GCG-Individual focuses on generating a tailored suffix designed specifically for a particular prompt. In contrast, GCG-Universal seeks to identify a generalized suffix that can be applied across multiple prompts, enabling it to deceive the model in a broader range of scenarios. 
\paragraph{AutoDAN} \cite{liu2024autodan} utilize a meticulously designed hierarchical genetic algorithm and generate stealthy jailbreak prompts. The generated prompts are highly readable and transferable.
\paragraph{PAIR} \cite{chao2023jailbreaking} is a jailbreak method that leverages an attacker LLM aiming at making the target LLM answer harmful prompts. The attacker LLM iteratively queries the target LLM to update and refine a candidate jailbreak prompt. 
\paragraph{DeepInception} \cite{li2023deepinception} proposes a simple method that uses the personification ability of LLMs. It creates a virtual and layered scene, allowing the model to find flexible ways to bypass usage controls in normal situations.
\subsection{Defense Methods}\label{defense_setting}
\paragraph{PPL} \cite{alon2023detecting} discovers that jailbreak prompts often lead to high perplexity values in LLMs. It, therefore, involves adding a classifier trained on perplexity and text length at the end of LLMs. The classifier can serve as a filter to avoid outputting potentially harmful answers.
\paragraph{Paraphrase} \cite{jain2023baseline} defense LLMs by making them paraphrase their inputs, avoiding deception caused by potential adversarial jailbreak suffixes within original inputs.
\paragraph{Retokenization} \cite{jain2023baseline} disrupts adversarial suffixes by retokenizing the input sequence, breaking tokens apart, and re-representing them with smaller tokens.
\paragraph{SafeDecoding} \cite{xu-etal-2024-safedecoding} is a safety-aware decoding strategy. It mitigates jailbreak attacks by amplifying the probabilities of safety disclaimers tokens among top-ranked tokens and attenuating the probabilities of harmful token sequences, ensuring LLMs generate helpful and harmless responses.
\paragraph{Self-Exam} \cite{phute2023llm} triggers the LLMs' awareness of safety issues by adding a pre-defined prompt template, asking LLMs to examine if their outputs are safe. 
\paragraph{Self-Reminder} \cite{xie2023defending} defends the LLMs by adding a system prompt specifically emphasizing potential safety issues to prevent the model from outputting harmful responses.
\textcolor{blue}{
\paragraph{Intention Analysis} \cite{zhang2025intention} enhances LLM safety by guiding the model to explicitly identify the essential intention of user queries and subsequently generate policy-aligned responses to mitigate harmful outputs.
}
\subsection{Metrics}\label{metrics}
\subsubsection{Calculating DSR}\label{cal_dsr}
We use Dic-Judge \cite{zou2023universal} to calculate DSR. Dic-Judge detects the presence of pre-defined refusal strings in the outputs of LLMs. Given a jailbreak query as input, if a particular refusal string is detected in the outputs, we consider the model successfully refuses to answer the question, thereby achieving the goal of defense; otherwise, the model is deemed to be deceived by the jailbreak inputs.
DSR is defined as follows:

\[\mathrm{DSR=\frac{|\{responses \ containing\ refusal\ strings \}|}{|\{input\ jailbreak\ queires\}|}}.\]

We adopt refusal strings as proposed by \citet{xu-etal-2024-safedecoding}, presented in Table \ref{tab:refusal_strings}.
\subsubsection{Calculating Overall Scores}\label{cal_overall}
We use Runtime per Query and overall averaged Just-Eval scores to calculate overall scores:
\[\mathrm{Overall=\big[1-Norm(t)+Norm(Avg.)\big]/2,}\]
where \(\mathrm{t}\) is Runtime per Query and \(\mathrm{Avg.}\) represents overall averaged Just-Eval scores. \textcolor{blue}{Here, \(\mathrm{Norm(x)}\) denotes the normalization of a metric \(\mathrm{x}\), calculated as \(\mathrm{x}\) divided by the sum of \(\mathrm{x}\) and the corresponding values for all reference groups.}

\begin{table}[h]
\centering
\begin{tabular}{l}
\toprule
\textbf{Refusal Strings} \\ \midrule
I'm sorry \\ 
Sorry \\ 
I am sorry \\ 
I apologize \\ 
As an \\ 
As an AI \\ 
I'm an \\ 
I'm just \\ 
As a language model \\ 
As an Assistant \\ 
I cannot \\ 
I do not \\ 
It is not</s> \\ 
It is not appropriate \\ 
I'm unable to \\ 
I am unable to \\ 
I am not allowed to \\ 
I am an AI language model \\ \midrule
\bottomrule
\end{tabular}
\caption{Refusal strings.}
\label{tab:refusal_strings}
\end{table}
\textcolor{blue}{
\section{Supplementary Experiments}\label{sup-main-exp}
\subsection{Scale Up Baselines and Data}
To further strengthen our evaluation, we expanded the jailbreak test samples from the initial 50-behavior subset to the full AdvBench (\~520 samples) and applied the jailbreak method from Section 3.2 for a more intuitive observation. The defense performances on Vicuna-7B-v1.3 are presented in Table~\ref{tab:scale-up-data-exp}.The results remain consistent with Table~\ref{tab:dsr_main}, reaffirming ABD’s strong defense capability against various stealthy jailbreak methods.
\subsection{Full Results on Qwen and Vicuna-13B}\label{more-models}
The results are shown in Table~\ref{tab:more-model-exp}.  
ABD remains performing decent on other sizes of models. Besides, in Vicuna-13B, 6 out of 9 selected layers are early and intermediate layers (layer 2, 11, 14, 19, 26, 29), and in Qwen, 2 out of 3 layers (layer 5, 14). This supports our interpretations of jailbreak mechanisms, as well as the conclusion that early and middle layers are most vulnerable to jailbreaks, basically hold in larger and smaller models.
}
\clearpage
\afterpage{
    \begin{figure*}[p]
        \centering
        \includegraphics[width=\textwidth,keepaspectratio]{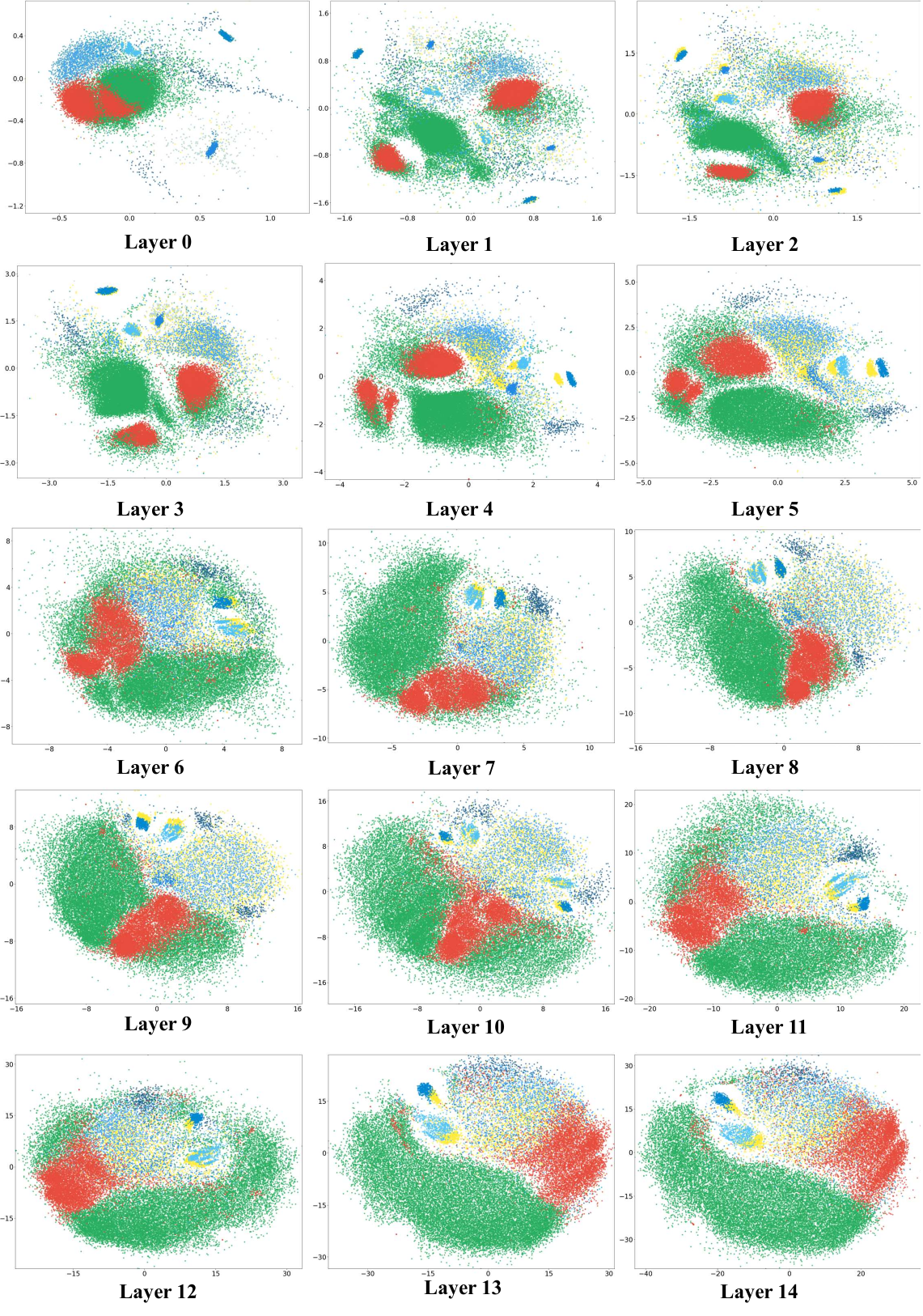}
        \caption{Activation spaces from layer 0 to layer 14.}
        \label{fig:all_subfig1}
    \end{figure*}
}
\afterpage{
    \begin{figure*}[p]
        \centering
        \includegraphics[width=\textwidth,keepaspectratio]{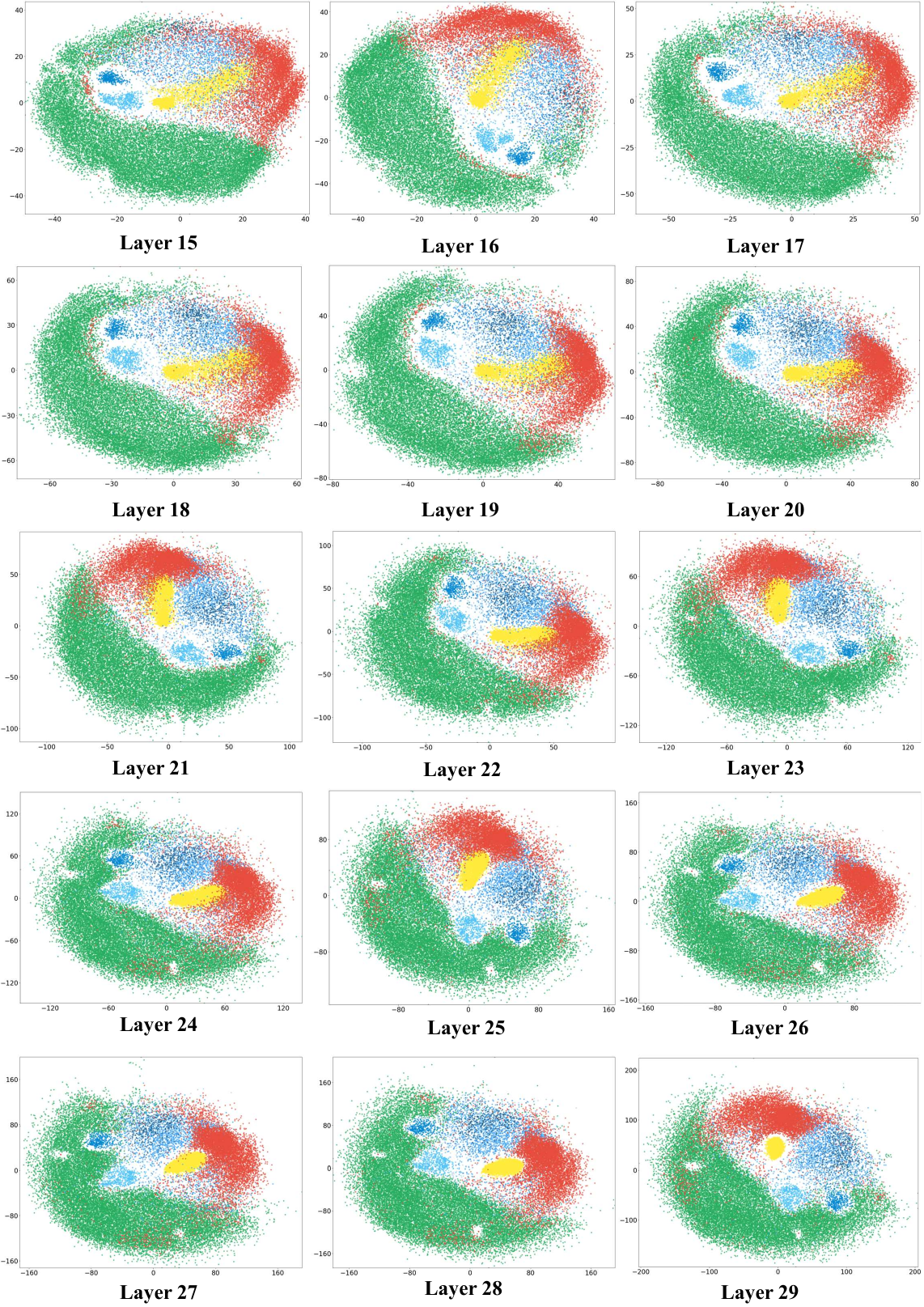}
        \caption{Activation spaces from layer 15 to layer 29.}
        \label{fig:all_subfig2}
    \end{figure*}
}
\clearpage
\afterpage{
    \begin{figure*}[t] 
        \includegraphics[width=\textwidth,keepaspectratio]{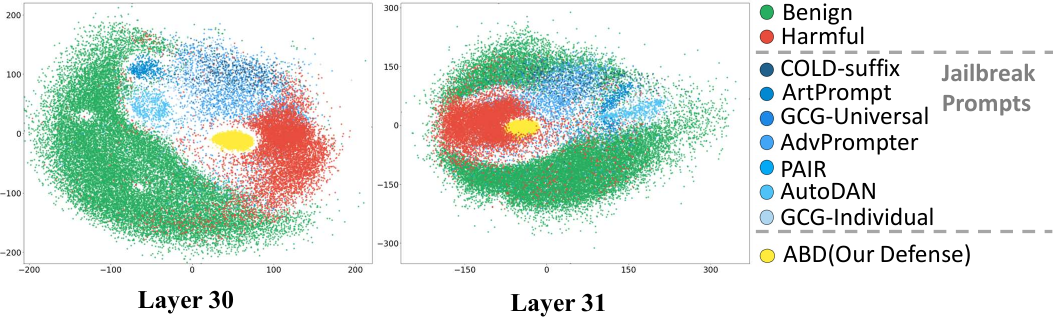}
        \caption{Activation spaces from layer 30 to layer 31.}
        \label{fig:all_subfig3}
    \end{figure*}
}
\end{document}